\documentclass{article} 
\usepackage[preprint]{colm2026_conference}

\usepackage{microtype}
\usepackage{hyperref}
\usepackage{url}
\usepackage{booktabs}
\usepackage{graphicx}
\usepackage[most]{tcolorbox}
\newtcolorbox{promptbox}{colback=gray!5,colframe=gray!60,boxrule=0.4pt}

\newcommand{\pair}[3]{%
  \begin{minipage}{0.48\textwidth}
    \centering
    \includegraphics[width=0.48\linewidth]{#1}%
    \hspace{0.1cm}
    \includegraphics[width=0.48\linewidth]{#2}\\
    \vspace{0.1cm}

     #3 
  \end{minipage}%
}
\usepackage[rightcaption]{sidecap} 
\usepackage{subcaption}

\usepackage{lineno}

\definecolor{darkblue}{rgb}{0, 0, 0.5}
\hypersetup{colorlinks=true, citecolor=darkblue, linkcolor=darkblue, urlcolor=darkblue}

\title{Multimodal Language Models \\ Cannot Spot Spatial Inconsistencies}

\author{Om Khangaonkar$^1$, Hadi J. Rad$^{2,3}$, Hamed Pirsiavash$^1$ \\
$^1$University of California, Davis\hspace{0.1cm} $^2$Shell \hspace{0.1cm} $^3$TU Delft\\
\href{https://reachomk.github.io/spatial-incon}{\nolinkurl{reachomk.github.io/spatial-incon}}} 

\begin{document}

\ifcolmsubmission
\linenumbers
\fi

\maketitle

\begin{abstract}

Spatial consistency is a fundamental property of the visual world and a key requirement for models that aim to understand physical reality. Despite recent advances, multimodal large language models (MLLMs) often struggle to reason about 3D geometry across multiple views. Rather than asking models to describe scene attributes, we introduce a more challenging task: given two views of the same scene, identify the object that violates 3D motion consistency. We propose a simple and scalable method for generating realistic, spatially inconsistent image pairs from multi-view scenes, enabling systematic evaluation of this capability. Our results show that state-of-the-art MLLMs significantly underperform human observers and exhibit substantial variability across different scene attributes, revealing a fragile and incomplete understanding of 3D structure. We hope our findings underscore the need for approaches that develop a more deeply grounded understanding of the physical world.

\end{abstract}

\section{Introduction}
\label{sec:intro}

Spatial consistency is a defining property of the physical world. As a camera moves through a static scene, objects change appearance in predictable ways governed by geometry, depth, and viewpoint. Their shapes, scales, and occlusions vary predictably and coherently across views. Humans are sensitive to violations of these relationships: any configuration that cannot physically exist immediately feels “off”. For any model that aims to understand or synthesize realistic scenes or videos, it is essential to be able to spot when views are \textit{physically impossible} to reconcile as observations of a unique static scene.


We hypothesize that while multimodal large language models (MLLMs) can describe individual views with impressive detail and accuracy, they fail to recognize where two views of a scene are mutually inconsistent. If this is true, it suggests today’s SOTA MLLMs demonstrate only surface-level 3D visual understanding and lack a robust internal model of spatial geometry. As MLLMs are increasingly used to percieve and judge the physical world, this limitation increasingly becomes a bottleneck.

To study this problem systematically, we focus on a simple but important question: \emph{can a model spot a spatial inconsistency between two images of the same scene?} This task directly probes whether a model understands 3D structure rather than merely memorizing visual statistics. A key challenge, however, is obtaining high-quality pairs of images that differ only by a physically inconsistent object. Existing 3D rendering or view-synthesis methods, such as Zero123 \citep{liu2023zero1to3} or SEVA \citep{zhou2025stable}, can generate alternative viewpoints, but are computationally expensive and often hallucinate unrealistic geometry that confound evaluation. We instead develop a simple, lightweight, fully automatic procedure that uses real multi-view scenes to create natural yet inconsistent image pairs. The algorithm requires no manual annotation and operates at more than 5 pairs per second on a single 3090, serving as a scalable tool to enable this study and future works.
\begin{figure*}[!t]
    \centering
    \includegraphics[width=0.99\textwidth]{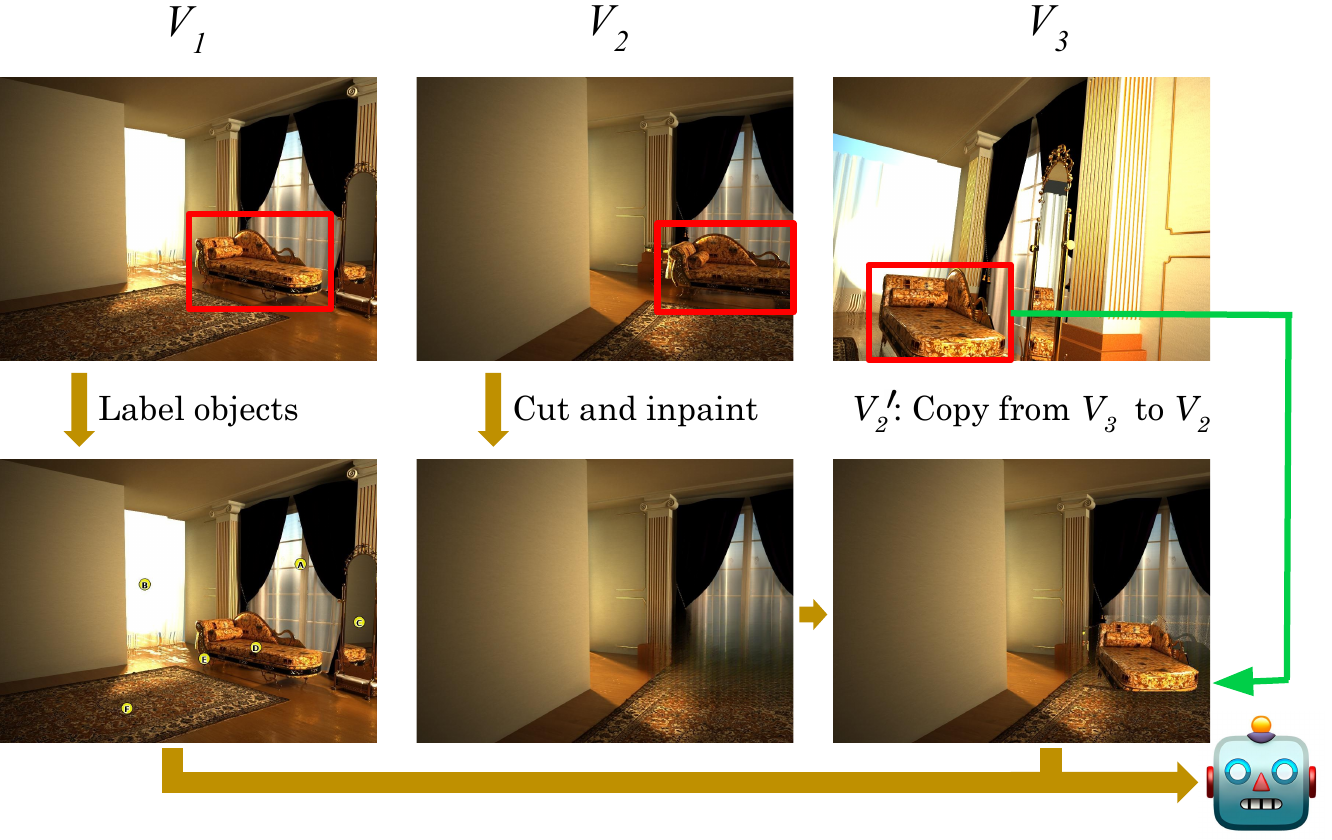}
 \vspace{-.1in}
\caption{
    \textbf{Synthesizing spatially inconsistent image pairs from multi-view data.}
    Given three views $(V_1,V_2,V_3)$ of the same static scene, we (1) select an object $O$ visible in all views, (2) Erase $O$ in view $V_2$ and inpaint to obtain a clean background, and (3) paste the instance of $O$ from view $V_3$ back into $V_2$ at its original location. Because $V_3$ is captured from a different camera pose, the pasted object has an appearance that is incompatible with the 3D geometry implied by $V_1 \rightarrow V_2$, while all other objects remain consistent. This yields realistic image pairs with controlled spatial violations and no manual annotation. We label objects in $V_1$ for our forced-choice evaluation. 
    }
    \label{fig:cut-paste}
\end{figure*} 
Given three views of the same static scene, we select the first as a reference, segment an object from the second and third views, and perform three steps: (1) remove the selected object from the second view and inpaint the missing region using an off-the-shelf inpainting model, (2) copy and paste the segmented object from the third view into the second, and (3) form a pair consisting of the first (reference) and the second (modified) view. Because the camera poses differ across the three views, the pasted object naturally acquires a 3D pose inconsistent with the reference view. The resulting pair preserves photorealism while introducing a controlled geometric violation, precisely the type of inconsistency that humans can easily detect but confuses most recent MLLMs.

Using this method, we construct a dataset of spatially inconsistent image pairs. Each pair includes ground truth annotations specifying which object violates consistency. We formulate a simple evaluation task: given two images of a scene, identify the inconsistent object. This task serves as a direct test of spatial reasoning ability in both humans and MLLMs. We evaluate several state-of-the-art models, including \textsc{GPT-5}, \textsc{Gemini 2.5 Pro}, and \textsc{Qwen3-VL}, and find that human participants achieve significantly higher accuracy than all tested models.



More importantly, these failures are not merely a matter of lower average accuracy. We find that models' correct answers are highly unstable across scene attributes, unlike humans', which are very robust to these changes. For example, while human performance remains comparatively stable across viewing conditions, MLLM performance fluctuates sharply with factors such as object depth, lighting, physical plausibility, and object or scene category. We also find that while models often fail on similar pairs, the incorrect answers to these pairs vary widely, suggesting divergence in spatial reasoning across models. This suggests that current MLLMs lack a robust and human-like representation of spatial structure. These results demonstrate that our spatially inconsistent pairs provide a challenging and informative benchmark for diagnosing spatial reasoning gaps in current multimodal systems.

Thus, we present a simple, scalable algorithm to automatically generate spatially inconsistent image pairs, a dataset and evaluation task that reveal clear gaps between human and model reasoning, and an in-depth investigation of several factors that may effect their performance. We hope our findings inspire future work to build in spatial inconsistency identification as a core capability for the next generation of MLLMs.  

\section{Related Works}
\label{sec:rw}
\subsection{Evaluating Perception in Multimodal Language Models}
Since the early days of VQA, many have questioned the perceptual limits of multimodal models. For example, the CLEVR dataset \citep{johnson2017clevr} was introduced as a diagnostic to test whether models were learning superficial statistics or genuinely understanding low-level spatial and compositional relationships. This concern has re-emerged with general-purpose MLLMs, which excel at semantic tasks but often fail at core perception. A crucial line of work, exemplified by BLINK \citep{fu2024blink}, found that many MLLMs ``can see but not perceive.'' They showed that while models can solve high-level VQA, they fail at basic vision tasks like correspondence or object localization that cannot be solved with captions alone. These findings spurred a new generation of more rigorous benchmarks.

To probe a model's intuitive ``world model,'' benchmarks like Physion++ \citep{tung2023physion++} and PhysBench \citep{chow2025physbench} evaluate understanding of physical properties and dynamics. Similarly, to test reasoning across multiple images, benchmarks such as MVBench \citep{li2024mvbench}, MMRB \citep{cheng2025evaluating}, and MM-Spatial \citep{mm-spatial} evaluate a model's ability to aggregate spatial and semantic information from different viewpoints. More recently, works such as 3DSRBench \citep{ma20253dsrbench} and 11Plus-Bench \citep{li202511plus} further study spatial reasoning in multimodal models, introducing diagnostic tasks for evaluating geometric relationships and spatial reasoning capabilities across diverse visual contexts. However, these works ask models to describe physical scene properties (e.g., stability, mass) or temporal details (e.g., ``what is the person doing after cleaning?''), while we isolate and evaluate a model's understanding of \textit{projective geometry}. Additionally, we test not \textit{what} is in a scene, but whether the scene's geometry is \textit{fundamentally possible}.
\begin{figure*}
  \centering
  \begin{tabular}{c}

    \pair{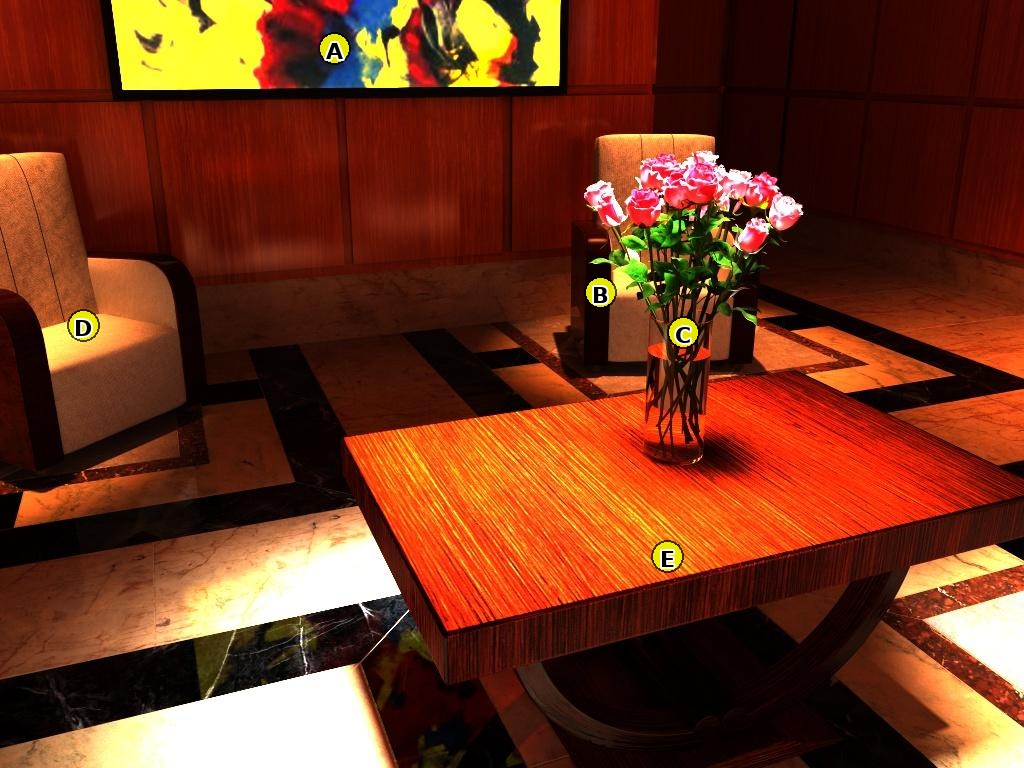}
         {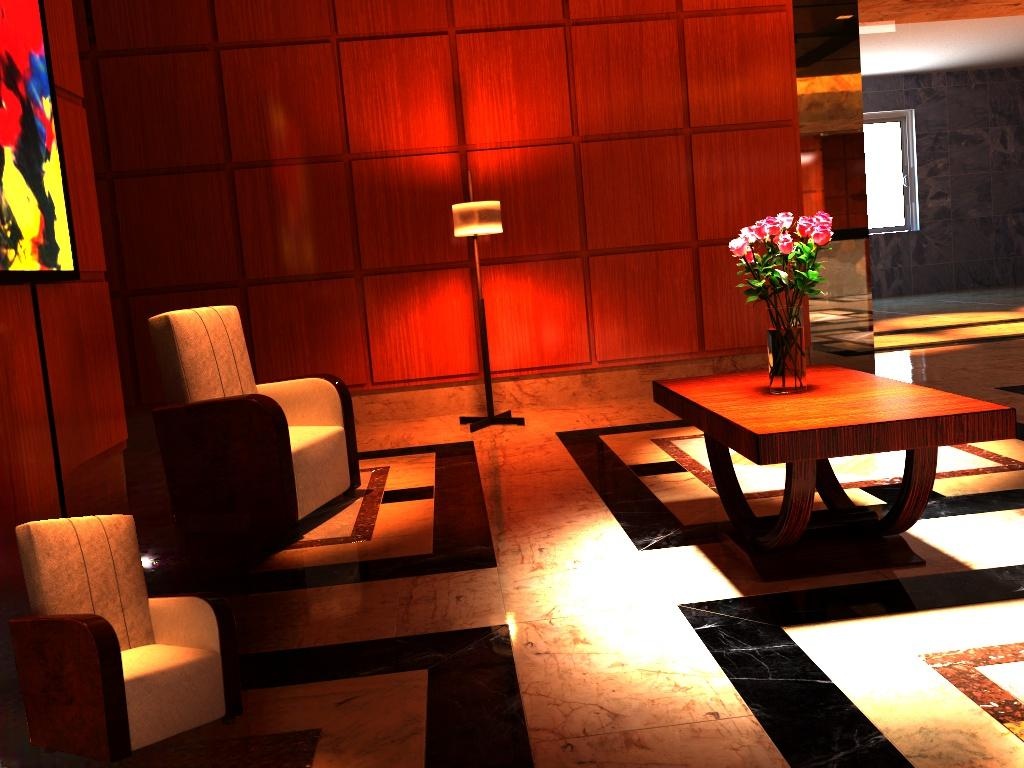}
         {\scriptsize \textbf{Depth:} Med, \textbf{Light:} High, \textbf{Implausible}, \textbf{\# of Labels:} 5, \\ \textbf{Object:} Chair, \textbf{Scene:} Lounge, \textbf{Correct:} D, \textbf{LLM:} C, D, B} 

    \pair{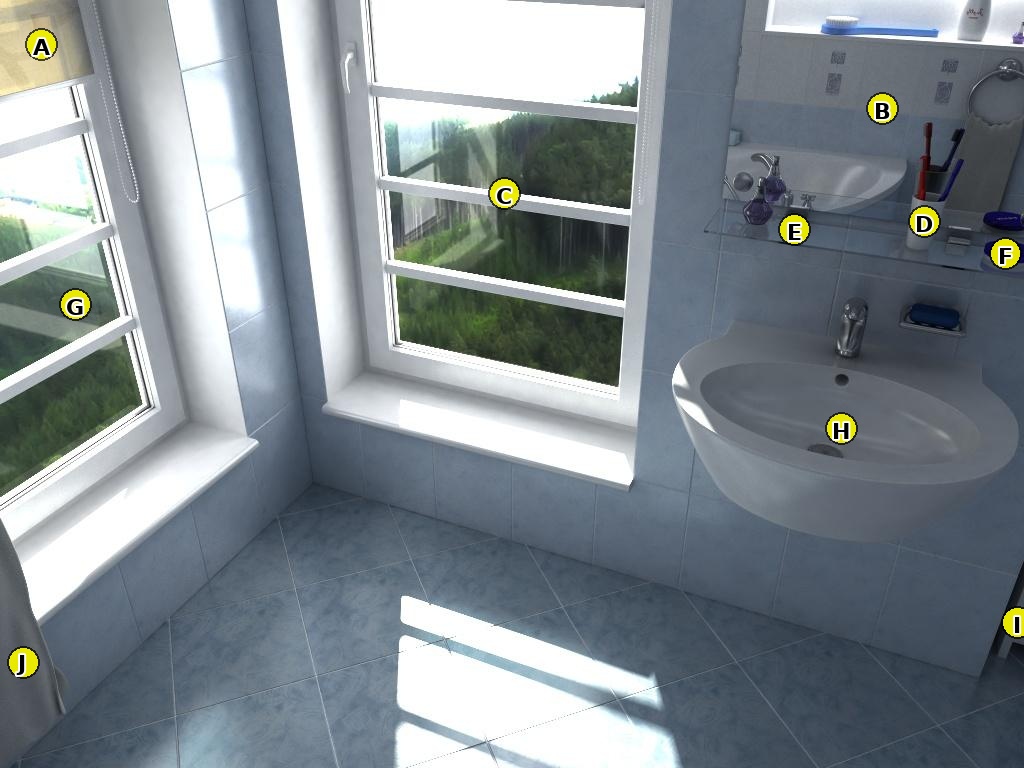}
         {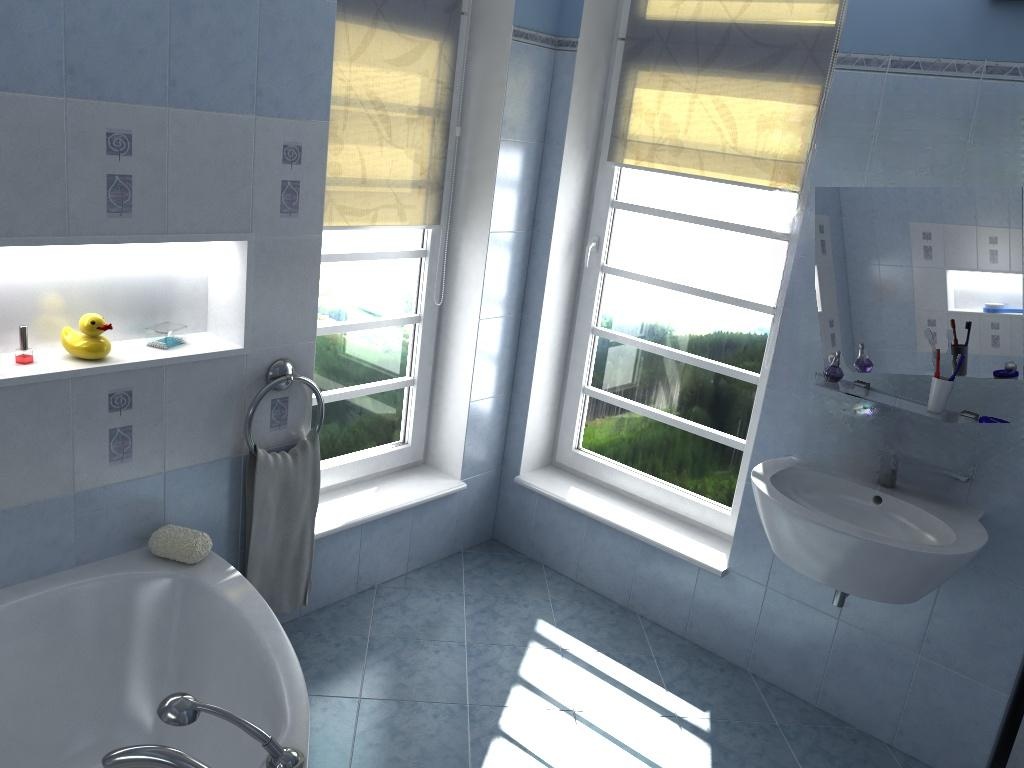}
         {\scriptsize \textbf{Depth:} Med, \textbf{Light:} Med, \textbf{Implausible}, \textbf{\# of Labels:} 10, \\ \textbf{Object:} Mirror,  \textbf{Scene:} Bathroom, \textbf{Correct:} B, \textbf{LLM:} A, J, J} \vspace{0.2cm}\\

    \pair{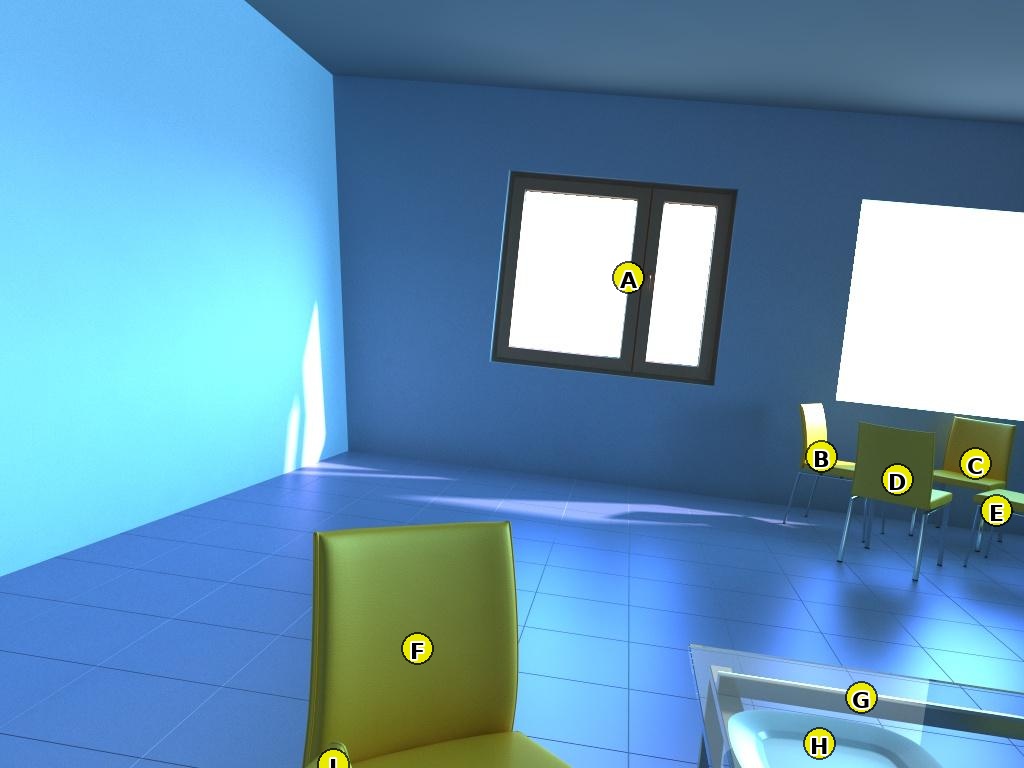}
         {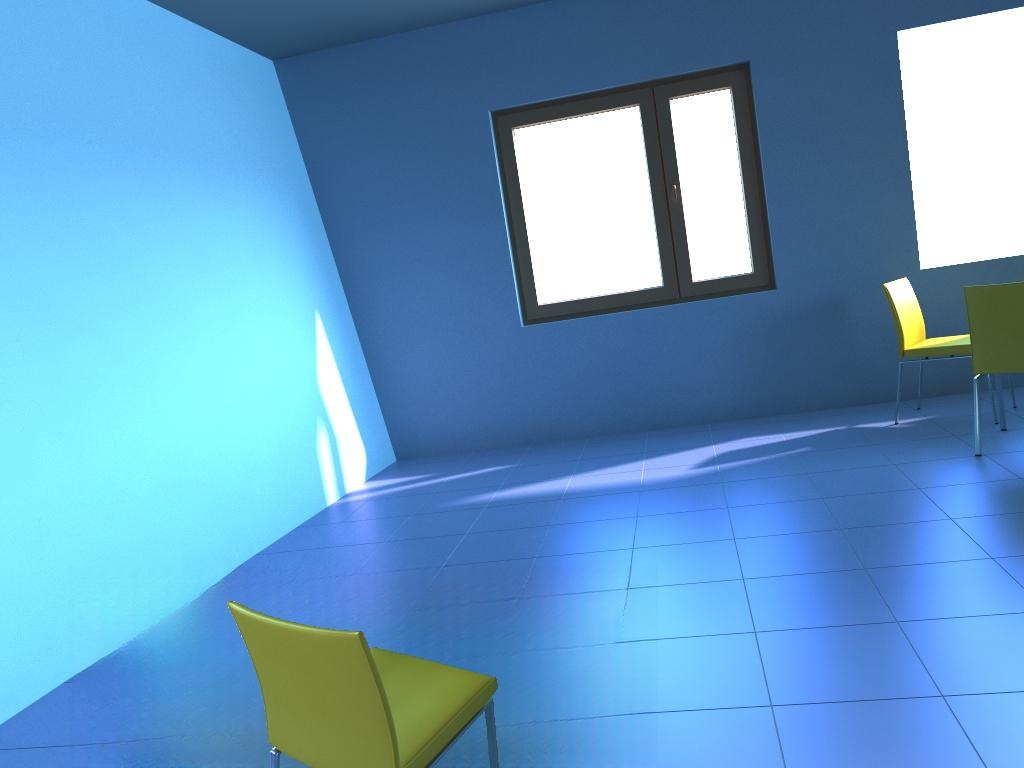}
         {\scriptsize \textbf{Depth:} Low, \textbf{Light:} Med, \textbf{Implausible}, \textbf{\# of Labels:} 9, \\ \textbf{Object:} Chair, \textbf{Scene:} Misc, \textbf{Correct:} F, \textbf{LLM:} F, F, F} 
    \pair{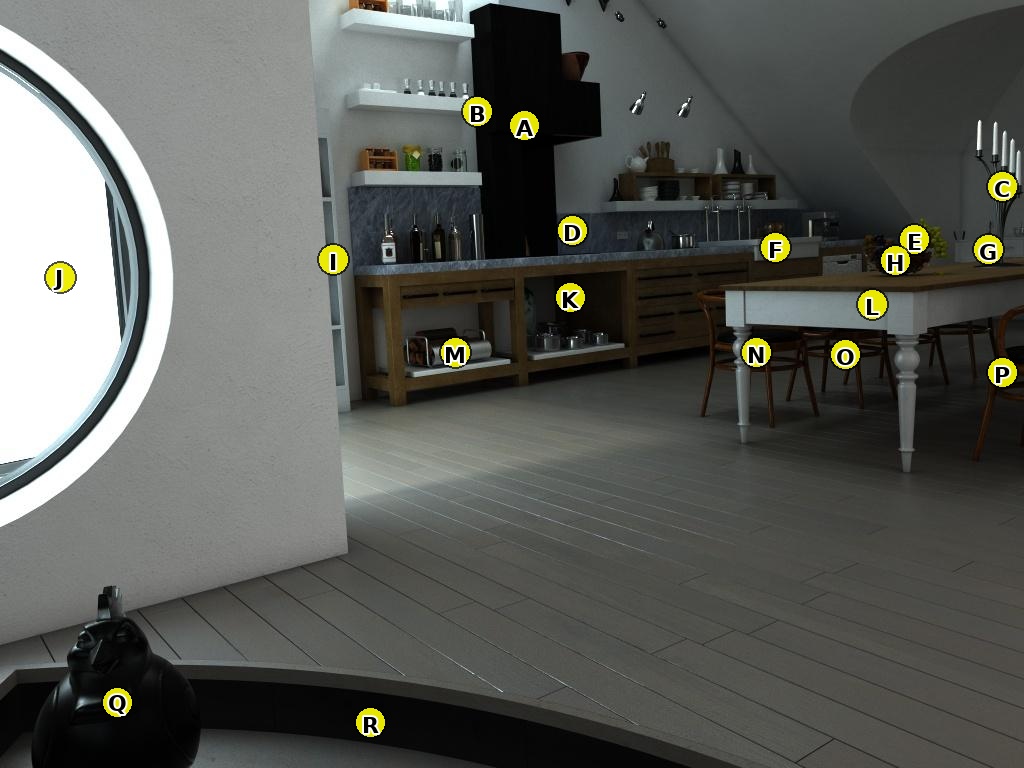}
    {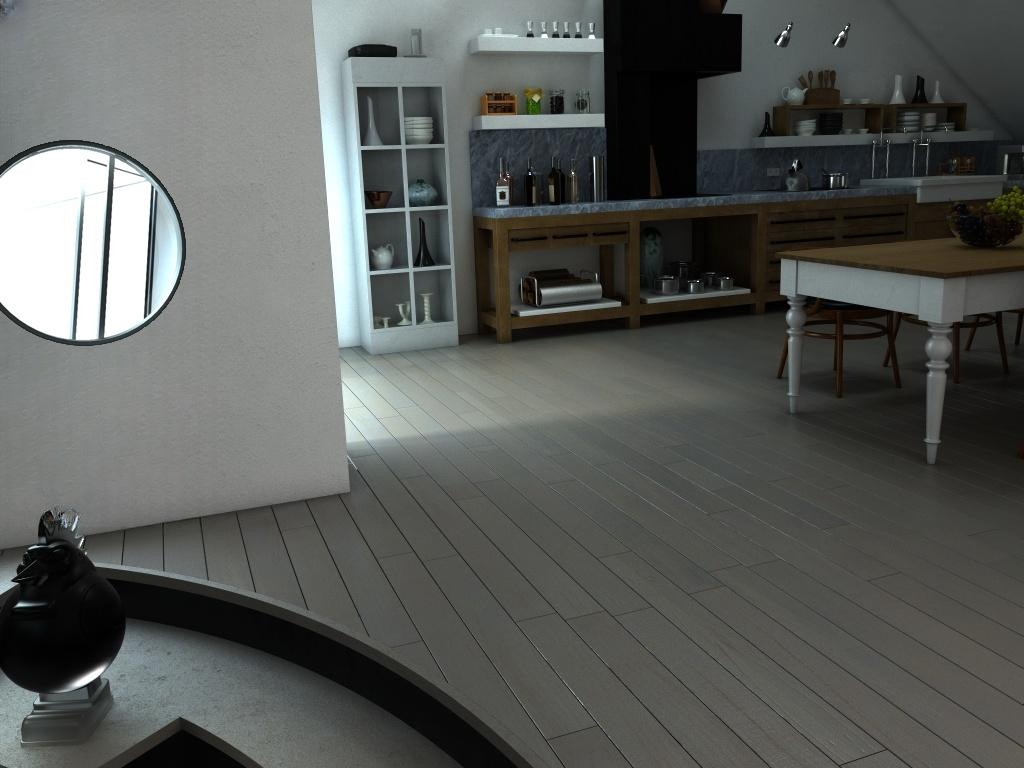}
    {\scriptsize \textbf{Depth:} High, \textbf{Light:} Med, \textbf{Plausible}, \textbf{\# of Labels:} 18, \\ \textbf{Object:} Window, \textbf{Scene:} Kitchen, \textbf{Correct:} J, \textbf{LLM:} E, Q, J} \\
  \end{tabular}
  \caption{\textbf{Example spatial inconsistencies. } AI labels are ordered as \textsc{GPT-5 (LR)}, \textsc{Gemini 2.5 Pro (MR)} and \textsc{Qwen3-VL 8B Instruct}. Zoom in to see labels in detail.}
\end{figure*}

\subsection{Counterfactual Scene Understanding}

A robust visual system should not only recognize what is present but also reject what is impossible, and counterfactual evaluation has become crucial to identify flaws in MLLMs visual representations. Early foundational works in this domain, such as Winoground \citep{thrush2022winoground} utilize semantic counterfactuals (e.g., swapping object-attribute bindings or word order) to demonstrate that models often ignore syntax and relational logic in favor of bag-of-words matching. This line of inquiry has since expanded to ``hard negatives,'' with benchmarks like CREPE \citep{ma2023crepe} and EqBen \citep{wang2023equivariant} testing a model's ability to distinguish between minimally different captions or images.

More recently, this paradigm has shifted toward visual and physical counterfactuals, where images are manipulated to contradict physical laws rather than just textual descriptions. HallusionBench \citep{guan2024hallusionbench} introduced a suite of visual illusions and consistency checks to probe whether MLLMs effectively hallucinate details when faced with confusing visual inputs. 
In the video domain, VideoHallu \citep{li2025videohallu} and Vinoground \citep{zhang2024vinoground} generate synthetic videos that violate physical principles (e.g., gravity, object permanence) or temporal logic, identifying a critical gap in dynamic causal reasoning. Similarly, Visual Jenga \citep{bhattad2025visual} introduces a new task to evaluate a generative model's notion of scene stability and object dependence by asking the model to progressively remove objects from the scene without deforming it or creating a physical inconsistency. Recent works have also begun exploring counterfactual diagnostics for spatial and physical reasoning in multimodal systems, highlighting persistent failures when models must reason about physically grounded scene structure rather than semantic relationships alone.

However, these existing counterfactual benchmarks largely focus on semantic plausibility (is the correct object present?) or dynamic plausibility (did it move correctly over time?). We instead create multiview counterfactuals that violate the 3D motion of the physical world. 
\vspace{-0cm}

\section{Synthesizing Spatial Inconsistencies}
\label{sec:method}

\subsection{Problem Setting}

Our goal is to construct pairs of views of a static scene in which a single object violates the spatial constraints implied by camera motion. Given multi-view imagery of a scene, we aim to generate two images that differ only in the 3D-consistency of one object. While one could synthesize such pairs using novel-view generation models (e.g., Zero123-style approaches \citep{liu2023zero1to3, zhou2025stable}), these methods are often slow, may introduce unintended artifacts, and may embed model-specific biases into the benchmark. 

Instead, we adopt a lightweight cut-and-paste strategy that operates directly on observed views. This approach scales to large datasets, runs in almost 5 pairs per second on a single GPU, and enables us to operate on challenging compositional scenes. Most importantly, by sourcing the inserted object from a true novel view, our method produces a controlled 3D violation that alters only the object's pose without altering its structure or texture. The resulting pairs are photorealistic and contain a single, well-defined spatial inconsistency.

\subsection{Source Data and Triplet Selection}

We build on the Hypersim dataset \citep{roberts:2021}, which provides multi-view indoor scenes with tracked instance-level masks. We select Hypersim because of its large scale nature with many scenes from diverse categories. Additionally, because the scenes are selected for their hyper-realistic nature, the images appear very close to real-world scenes in their detail and complexity while giving us access to pixel-perfect annotations of segmentation, depth, intrinsic images, and more. However, as these properties can be inferred from other models, we can also extend our method to real-world multiview scenes, which we explore in the Appendix using DL3DV \citep{ling2024dl3dv}.

For each scene, we sample three frames $(V_1, V_2, V_3)$ and select an object $O$ visible in all three views. To avoid degenerate cases where the views are nearly identical, we impose that at most $75\%$ of the objects overlap between $(V_1,V_2)$ and between $(V_2,V_3)$. We further restrict selection of $O$ to objects occupying between $5\%$ and $10\%$ of the image area to ensure sufficient visual detail while avoiding large objects that lead to inpainting artifacts or small objects that lack textural cues. Additionally, we enforce the pixels of $O$ in $V_3$ must project to at 40\% of the area of $O$ in $V_2$, ensuring that $V_2$ and $V_3$ have enough overlap. 

\subsection{Violating Spatial Consistency}

Given a selected triplet $(V_1, V_2, V_3, O)$, we construct an edited version $V_2'$ in which only object $O$ violates the geometric constraints consistent with the transition of camera view point $V_1 \!\to\! V_2$. The process consists of two steps:

\begin{enumerate}
    \item \textbf{Removing $O$ in $V_2$.}  
    We erase $O$ from view $V_2$ by inpainting its bounding box (enlarged by $20\%$ to include context) using the LaMa inpainting model \citep{suvorov2021resolution}. This produces an object-free background that remains visually coherent. We inpaint the bounding box instead of the segmentation mask to ensure contour details of the object are also erased.
    
    \item \textbf{Reinserting $O$ from $V_3$.}  
    We extract $O$ from $V_3$ using its instance segmentation mask and paste it into the inpainted region of $V_2'$. The pasted object preserves its aspect ratio and is centered to match the size of $O$ in $V_2$. All pixels belonging to other object instances within the inpainted area are restored to maintain continuity and correct occlusion ordering.
\end{enumerate}

\subsection{Benchmark Construction}

 We first auto-generate over 10,000 pairs. Following established practice to validate the benchmark \citep{fu2024blink}, we manually inspect all pairs. Because of the time-consuming nature of manual filtering, we select 1-2 pairs per scene resulting in a set of 615 pairs. While this value is similar to prior works such as BLINK \citep{fu2024blink}, for a larger-scale evaluation, one can use our whole auto-generated set. To probe  MLLMs' spatial reasoning, we convert each pair into a forced-choice question: given views $(V_1, V_2')$, identify the object that violates 3D consistency.

\begin{table*}[t]
\centering
\small
\begin{tabular}{l c | l c}
\toprule
\textbf{Model} & \textbf{Overall} & \textbf{Model} & \textbf{Overall} \\
\midrule
\textsc{Gemma 3 12B}              & 8.5  & Random Chance            & 7.9  \\
\textsc{Idefics3 8B}              & 8.9  & Human                    & 84.8 \\ \cmidrule(lr){3-4}
\textsc{Idefics2 8B}              & 11.9 & \textsc{GPT-5 Nano}               & 15.3 \\
\textsc{Qwen2.5-VL 7B}            & 15.9 & \textsc{Gemini 2.5 Flash}         & 17.6 \\
\textsc{InternVL 3.5 8B}          & 16.1 & \textsc{GPT-4o}                   & 19.0 \\
\textsc{LLaVA OneVision 1.5 8B}   & 16.6 & \textsc{Gemini 2.5 Pro} (\textsc{HR})      & 28.9 \\
\textsc{SpaceQwen2.5-VL 3B}       & 18.2 & \textsc{Gemini 2.5 Pro} (\textsc{LR})      & 29.4 \\
\textsc{Llama 3.2 Multimodal 11B} & 23.4 & \textsc{Gemini 2.5 Pro} (\textsc{MR})      & 29.4 \\
\textsc{Qwen3-VL 4B}              & 24.7 & \textsc{GPT-5} (\textsc{HR})               & 30.2 \\
\textsc{Qwen3-VL 8B Thinking}     & 25.2 & \textsc{GPT-5} (\textsc{MR})               & 31.4 \\
\textsc{Qwen3-VL 8B Instruct}     & 27.6 & \textsc{GPT-5} (\textsc{LR})               & \textbf{34.2} \\
\cmidrule(lr){1-2}\cmidrule(lr){3-4}
Ensemble (open source)   & 30.1 & Ensemble (all models)    & 35.0 \\
\bottomrule
\end{tabular}
\caption{
\textbf{Overall accuracy (\%) on the spatial inconsistency identification task.}
We report overall accuracy for a random-choice baseline, humans, and a range of multimodal language models.
We find that humans substantially outperform all tested models. Interestingly, higher reasoning budget does not necessarily yield higher accuracy.
Ensembling open source models also allows them to perform at the level of large-scale proprietary models.
}
\vspace{-0.5cm}
\label{tab:main}
\end{table*}

\noindent  \textbf{Object labeling.}  
We label all objects in view $V_1$ larger than $1000$ pixels (assuming $1024 \times 768$ image size) with letter labels \texttt{A\dots Z} following prior work \citep{fu2024blink}. This labeling scheme allows models to refer to objects unambiguously in natural language. If more than $26$ objects satisfy the size threshold, we keep the largest $26$ while ensuring that $O$ is included. If too few objects qualify, we slightly relax the threshold but never include instances smaller than $300$ pixels that lack meaningful visual detail. These labels define the answer space for both human and model evaluation.

\noindent\textbf{Prompt and evaluation.}
At the test time, the model receives the labeled $V_1$ and the edited $V_2'$ along with the prompt, which is provided in the appendix. We parse the final output and mark the item correct if the letter matches the single edited object. Our primary metric is the average accuracy across pairs.


\noindent  \textbf{Human evaluation protocol.}
We evaluate both humans and multimodal language models under the same forced-choice protocol. Given an image pair, participants and models must indicate which letter marks the 3D-inconsistent object. No additional textual hints about the manipulated object or category are provided. We measure accuracy over all 615 pairs. Section~\ref{sec:results} reports detailed results broken down by depth, lighting, and physical plausibility.

\section{Experiments}
\subsection{Models Evaluated}
We evaluate a comprehensive suite of multimodal large language models (MLLMs) spanning various architectures, sizes, and training strategies. Our selection includes leading proprietary frontier models, specifically the \textsc{GPT-5} family, \textsc{GPT-4o}, and \textsc{Gemini 2.5} (Pro and Flash) under a wide range of reasoning budgets. We also assess state-of-the-art open-weight models, including \textsc{Llama 3.2 Multimodal} \citep{grattafiori2024llama3herdmodels}, \textsc{Qwen 2.5}, \textsc{Qwen3-VL Instruct}, and \textsc{Qwen3-VL Thinking} \citep{bai2025qwen25vltechnicalreport, Qwen3-VL}, \textsc{Gemma 3} \citep{team2025gemma}, \textsc{Idefics2 and 3} \citep{laurençon2024mattersbuildingvisionlanguagemodels}, \textsc{LLaVA OneVision 1.5} \citep{li2024llava}, and \textsc{InternVL 3.5} \citep{wang2025internvl3}. Finally, to investigate the impact of domain-specific fine-tuning and inference-time reasoning, we include spatially specialized variants such as \textsc{SpaceQwen2.5} \citep{chen2024spatialvlm}. For \textsc{GPT-5} and \textsc{Gemini 2.5 Pro}, we also report low (LR), medium (MR) and high (HR) reasoning levels. 

\subsection{How Many Spatial Inconsistencies Can MLLMs Catch?}
\label{sec:results}
 As shown in Table \ref{tab:main}, humans can easily spot spatial inconsistencies with relative ease at 84.8\% accuracy. However, most models struggle with our task, with the very best (\textsc{GPT-5} with ``low'' effort reasoning) achieves only 34.1\%. This suggests that learning to spot inconsistencies still remains a perceptual gap in multimodal language models, even as accuracy in many 3D tasks such as depth or correspondence approaches human capabilities \citep{bai2025qwen25vltechnicalreport}. Surprisingly, models such as \textsc{SpaceQwen}, which have been finetuned to excel at spatial reasoning tasks, still fail to spot inconsistencies. This may suggest that existing methods to teach models spatial understanding still leave large gaps in their 3D reasoning capabilities. Additionally, as is shown by the result corresponding to \textsc{Qwen3-VL}, \textsc{Gemini 2.5 Pro}, and \textsc{GPT-5}, increasing test-time compute \textit{does not} lead to improved performance. In fact, we consistently find that increasing reasoning effort often degrades model performance or leads to no further improvement. This finding is counterintuitive to the promise of visual reasoning and merits further investigation, which is out of the scope for this paper. 

Because the camera poses differ across $(V_1, V_2, V_3)$, the appearance of $O$ as seen from $V_3$ will generally not be geometrically compatible with its appearance in $V_1$. This edit therefore introduces a precise 3D inconsistency while leaving all remaining objects unchanged. Models with strong 3D geometric understanding must detect mismatches in scale, orientation, shading, correspondence, and occlusion patterns across the two views.

 We also report results after ensembling the models together. We assign each model votes based on its performance relative to the lowest performing model, \textsc{Gemma 3 12B}. We weight $i$'th model's binary vote using $\frac{\text{accuracy of model  } i}{\text{accuracy of \textsc{Gemma 3 12B}}}$, then, choose the answer with highest summation of weighted votes. As shown in Table \ref{tab:main}, we see a small but significant increase in performance when ensembling all models. Interestingly, ensembling just open-source models enables them to perform on par with proprietary models like \textsc{GPT-5} (HR).

\noindent  \textbf{Depth and lighting factors.}
Because Hypersim dataset provides ground-truth depth and per-pixel illumination, we can directly examine how geometric and photometric factors influence performance. For each inconsistent pair, we compute the depth of the modified object and the average scene brightness, then partition the dataset into 3 equally sized bins based on the empirical distributions: \emph{close}, \emph{medium}, and \emph{far} for object depth, and \emph{dark}, \emph{medium}, and \emph{bright} for lighting. For object depth, we use the ground-truth depth for frame $V_1$, and for lighting we use the luminance averaged across $V_1$ and $V_2$. This categorization enables systematic comparisons of human and model accuracy under different viewing conditions. Surprisingly, we find that most multimodal models perform substantially better on objects that are \emph{far} relative to those that are \emph{close}, with differences exceeding 15 percentage points, whereas human accuracy is highest for \emph{medium}-distance objects and remains stable for \emph{close} and \emph{far} ones. This pattern suggests that current MLLMs exhibit geometric biases related to depth perception. Lighting reveals similar discrepancies: humans are consistently more accurate in \emph{bright} scenes, while models tend to perform best under \emph{medium} illumination, potentially reflecting biases inherited from training data.

\begin{figure}[t]
    \centering
    \includegraphics[width=1\linewidth]{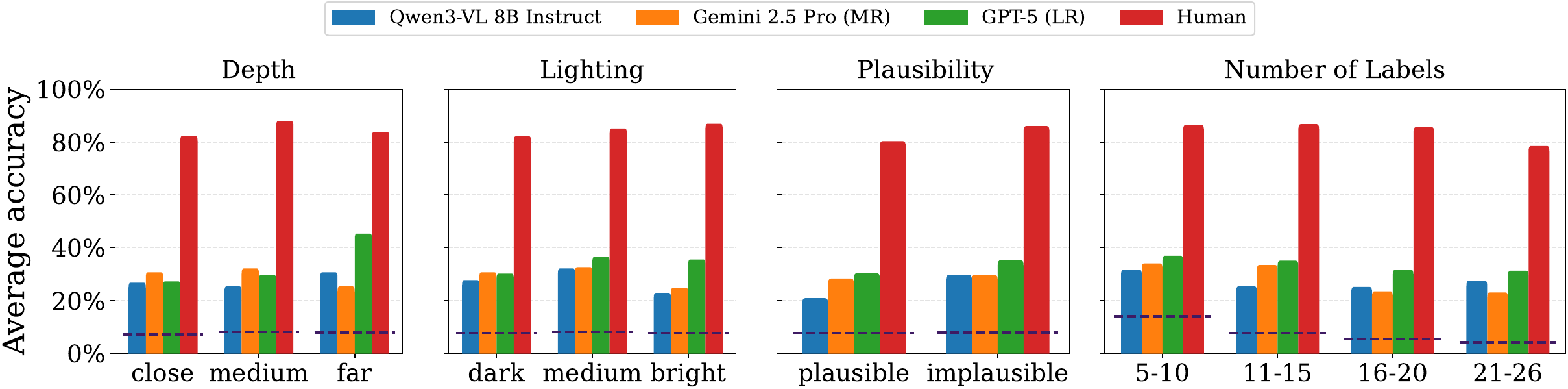}

    \caption{
    \textbf{Model accuracy varies across scene attributes, but much less across the number of labels per pair.}
    Left: We report accuracy for identifying the single spatially inconsistent object, stratified by inconsistent object depth (close/medium/far), average pair brightness (dark/medium/bright), and the augmented object's physical plausibility. While humans remain comparatively robust across conditions, models can show substantial sensitivity to changes in the pair's scene composition. Right: We partition the dataset based on the number of labels per pair and report accuracy. Interestingly, this has a minor effect on humans but not much of one on models. \textit{Dashed line represents random chance.}
    }
    \label{fig:depth_light_plaus_num_label_acc}
\end{figure}

\noindent  \textbf{Physical plausibility of poses.}
Not all spatially inconsistent poses are equally obvious. Some pose manipulations are obviously impossible in everyday scenes (e.g., a chair with two legs hovering above the ground), while others are subtle geometric violations that are harder to notice at a glance. To capture this variation, we categorize each pair by the \emph{physical plausibility} of the inconsistent pose. We estimate the camera transformation between frames $V_2$ and $V_3$ of the underlying Hypersim sequence and use the induced roll to approximate whether the inserted object appears stably supported or not. Pairs with less than $5^\circ$ roll are labeled as \emph{plausible}, meaning the object appears roughly stable under gravity, while those with larger roll are labeled as \emph{implausible}. This allows us to analyze how the physical compatibility of the inserted pose affects detection difficulty. We find that both humans and models more easily identify inconsistencies in implausible pairs, as these poses tend to be unusual in real-world settings. 

\noindent  \textbf{Number of labels per pair.} In Figure \ref{fig:depth_light_plaus_num_label_acc}, we group our evaluation set by the number of labels per pair and plot the accuracy across models and groups. Surprisingly, we do not see a major decrease in accuracy, even when the number of choices doubles or quadruples. This suggests that the model struggles to select the best choice out of a shortlist, and that adding more "choices" does not necessarily make the task harder.

\begin{figure}[t]
    \centering
    \includegraphics[width=1\linewidth]{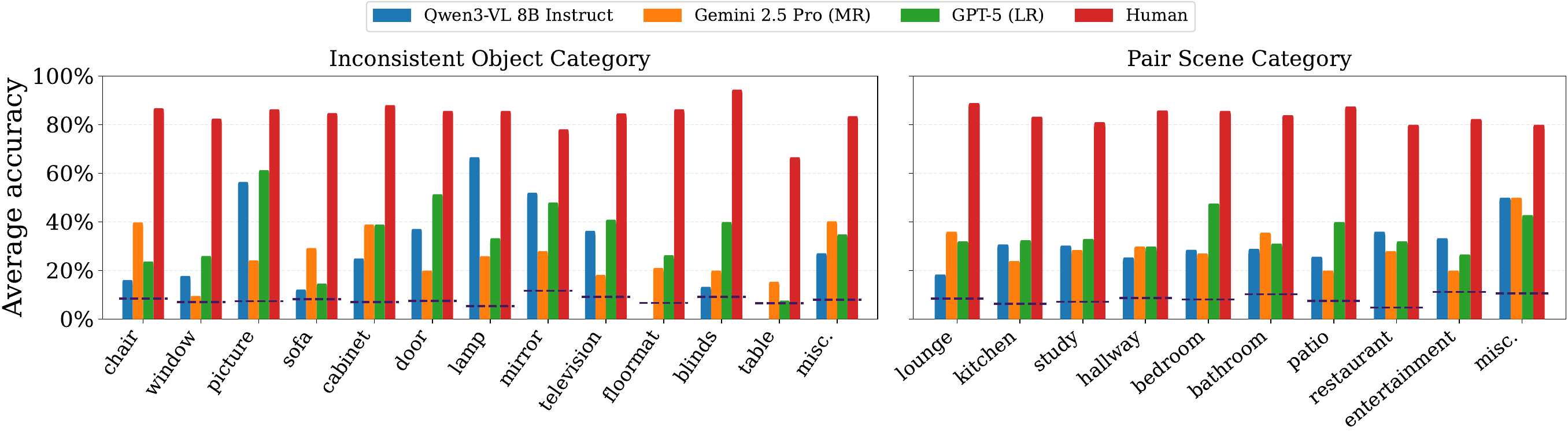}
    \vspace{-0.5cm}
    \caption{\textbf{Accuracy varies greatly across the inconsistent object or pair scene categories.} While humans are relatively consistent across all settings, the models show large amounts of variance. This suggests that their 3D understanding is brittle across our diverse visual world. \textit{Dashed line represents random chance.}  }
    \label{fig:object_scene_class}
     \includegraphics[width=0.95\linewidth]{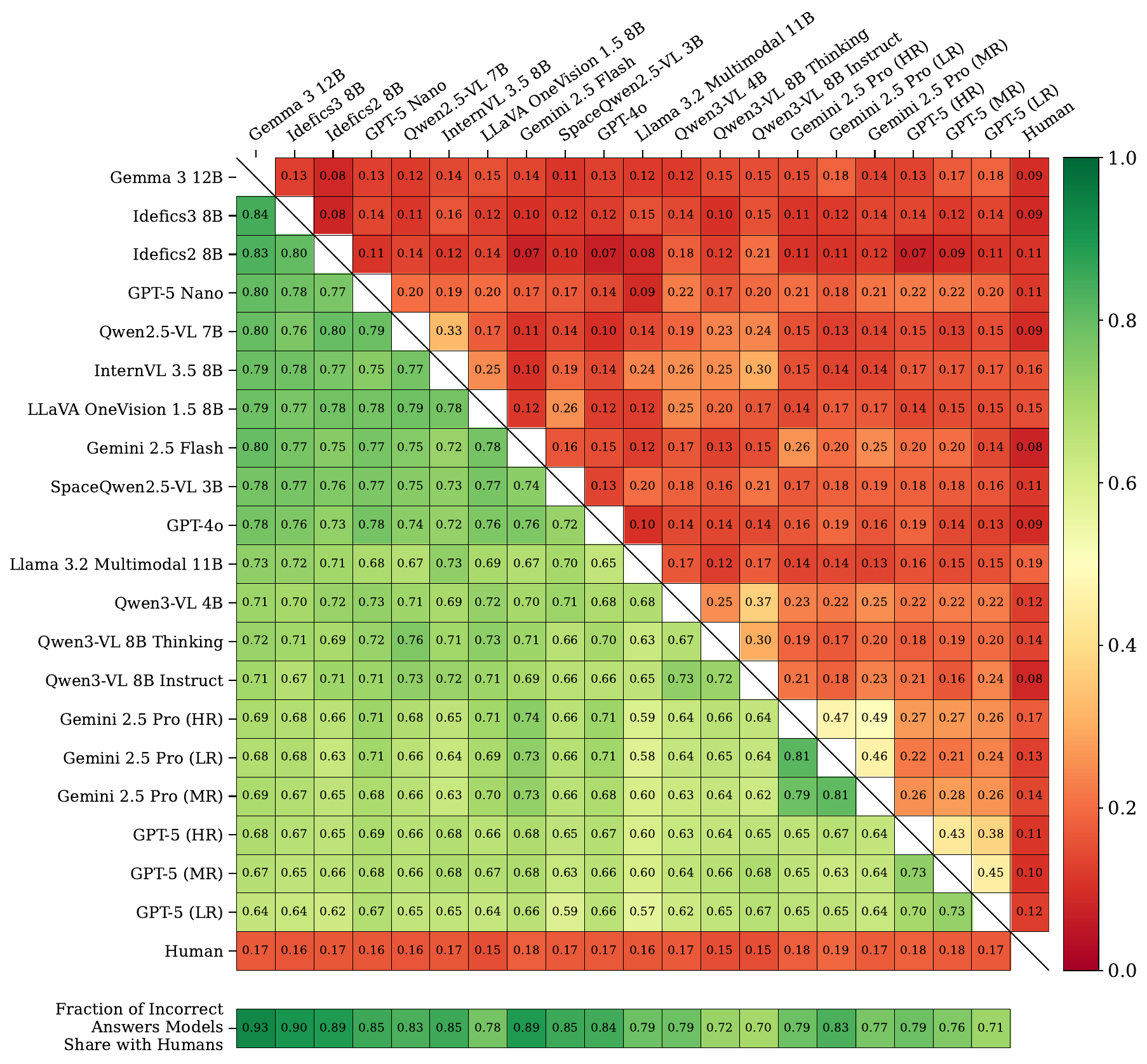}
    \caption{\textbf{Models get similar questions wrong, but don't pick similar answers. } On the bottom left, we report the IoU between two models' sets of incorrect questions. On the top right we take the intersection of incorrect questions between two models and report the fraction that they both produce the same wrong answer. From the last row, we find that most inconsistencies that trick the human, trick the AI models too.
    However, models rarely guess similar answers when wrong, suggesting large variance in 3D understanding.
    }
    \label{fig:same_wrong_heatmap}
    \vspace{-0.75cm}

\end{figure}
\begin{figure}[h]
    \centering


\end{figure}

\noindent  \textbf{Inconsistent object category. } We ask if models are equally strong at spotting inconsistencies across object types. We reuse the categories Hypersim annotates each object with. As some objects have an "unknown" category label, we group these into the "miscellaneous" category. We also add any pairs with inconsistent objects from categories with less than 10 pairs across the entire set into the "miscellaneous" category, as the individual sample size may be too small.



In Figure \ref{fig:object_scene_class}, we plot the accuracy across each class type for our top models and find that, while humans are consistent across object types (with each class's accuracy within a few points of the others), MLLMs accuracy varies greatly between classes. For example, while \textsc{Qwen3-VL 8B Instruct} has an average accuracy of 27.6, it ranges between 0 and 67\% depending on the category. We also see that model rankings vary greatly across object types, though none approach to human accuracy. For example, \textsc{Qwen3-VL 8B Instruct} surprisingly strongly outperforms \textsc{GPT-5 (LR)} and \textsc{Gemini 2.5 Pro (MR)}  at lamps.

\noindent  \textbf{Scene category. } To enable analysis across scene types, we group them into nine categories, plus a "miscellaneous" category. We first have \textsc{Qwen3-VL 8B Instruct} caption the scene for each pair's initial unmodified pairs $V_1$ and $V_2$, then feed all the captions into \textsc{GPT-5} and have it create and assign the 9 categories. We report results in Figure \ref{fig:object_scene_class}, where we observe a large amount of variance across categories. For example, \textsc{Qwen3-VL 8B Instruct} mostly outperforms \textsc{Gemini 2.5 Pro (MR)}, but gets lower overall accuracy due to its weak performance on Lounge scenes. Similarly, \textsc{GPT-5 (LR)}  gains most of its advantage from Bedrooms and Patios.

\noindent  \textbf{Overlapping incorrect answers. } Figure \ref{fig:same_wrong_heatmap} reveals a striking asymmetry between common wrong questions and common wrong answers. In the lower-left triangle, model–model pairs exhibit substantial overlap in the questions they answer incorrectly. In contrast, human–model IoU appears low, largely because humans make far fewer errors than the MLLMs, so the union is dominated by the model’s failures. To account for this imbalance, we add an extra row reporting the fraction of humans' wrong questions that are also wrong for the model. These results suggest that humans and models often get similar questions wrong. However, in the upper-right triangle, humans and the same models rarely produce the same incorrect answer for the same subset of failures that is common. Thus, tend to agree on which examples are hard but not on how they are wrong. This suggests that current 3D failures are not driven by a single coherent alternative interpretation of the scene, but instead reflect unstable and model-specific breakdowns in spatial reasoning.

\subsection{Are Models Spotting Inconsistencies or Artifacts?}
While our cut-and-paste method offers a simple and scalable way to synthesize inconsistencies, it operates in pixel space. This raises a natural question: are the models actually spotting 3D inconsistencies, or are we just evaluating artifact spotting? 

\begin{table} [h!]
  \centering
  \begin{tabular}{l|c|c|c}
    \toprule
     & {\scriptsize Our} & \scriptsize Artifacts only & \scriptsize No inconsistency or arti- \\
    \scriptsize Model & \scriptsize Benchmark & \scriptsize ($V_2$ self-paste) & \scriptsize facts ($V_2$ no change)  \\
    \midrule
    Random Chance & 7.9 & 7.9 & 7.9  \\
    \midrule
    \textsc{Qwen3-VL 4B} & 24.7 & 19.5 & 20.7 \\ 
    \textsc{Qwen3-VL 8B}  & 27.6 & 21.3 & 22.0  \\
    \textsc{Gemini 2.5 Pro  (MR)} & 29.4 & 13.8 & 15.3\\
    \textsc{GPT-5 (LR)} & 34.2 & 15.1 & 13.7 \\
    \bottomrule
  \end{tabular}

\caption{\textbf{Models do not succeed by detecting inpainting artifacts alone.} We compare performance on our full benchmark against two controls: a self-pasting baseline that preserves inpainting artifacts while removing the inconsistency, and a no-change baseline that contains neither artifacts nor inconsistency. Across all models, self-pasting performs similarly to the no-change baseline, indicating that artifact cues by themselves do not explain benchmark performance.}
  \label{artifact_merged}
\end{table}
We investigate this through two lenses. First, we design a "self-pasting" baseline that contains inpainting artifacts but \textit{no spatial inconsistency}. More specifically, given frames $V_1$, $V_2$, $V_3$, we remove the original object in $V_2$, expand and inpaint the bounding box and paste in the unmodified object from $V_2$ to $V_2$, instead of pasting a new view of the object from $V_3$. This allows us to ablate the effect of the artifacts in isolation from any spatial inconsistency. To contextualize our results, we compare against our original results (which have both artifacts and an inconsistency). Second, we use a "no change" baseline, which uses the unmodified $V_2$ image, and thus lacks both artifacts and an inconsistency. 

\begin{table} [t]
\centering
\small
\begin{tabular}{l|c|ccccc}
& \multicolumn{6}{c}{\# of self-pasted objects} \\

\midrule
Model & None & 1 & 3 & 5 & 10 & all \\
\midrule
\textsc{Qwen3-VL 4B Instruct} & 24.7 & 25.0 & 24.9 & 25.2 & 25.0 & 25.0 \\
\textsc{Qwen3-VL 8B Instruct} & 27.6 & 27.3 & 26.7 & 26.0 & 25.0 & 25.4 \\
\textsc{Gemini 2.5 Pro (MR)} & 29.4 & 28.1 & 29.8 & 28.9 & 26.7 & 24.7 \\
\textsc{GPT-5 (LR)} & 34.2 & 31.7 & 34.0 & 33.0 & 33.5 & 30.2 \\
\bottomrule

\end{tabular}

 \caption{\textbf{Self-pasting additional objects does not make our task harder. } Performance remains largely stable even when many or all objects are self-pasted, suggesting that models are looking for 3D structure and the task is not confounded by artifacts. }
 \label{tab:selfpaste-rand}
\end{table}

Interestingly, in Table \ref{artifact_merged}, "self-pasting" performs no better (and sometimes worse) than the "no change" baseline across all models. This suggests that inpainting artifacts are so small that they effectively go unnoticed by the model. Also, notice how the "no change" baseline performs higher than random chance, even though it lacks a spatial inconsistency. We hypothesize this is due to some common bias between  the model and our object selection criteria, and is likely similar to why accuracy stays mostly consistent as choices increase.

To further investigate the role of artifacts, we ask if introducing new artifacts outside of the inconsistency will make it more challenging for the model to select the modified object. To evaluate this, we self-paste several additional objects (ranging just one to all objects) in the scene to artificially synthesize inpainting artifacts. If our model did not have any 3D understanding and was simply detecting artifacts, it would be unable to select the inconsistency due to the amount of confounding artifacts. However, in Table \ref{tab:selfpaste-rand} we find that regardless of how many extra objects we self-paste, there is little to no effect on inconsistency detection. Even in the extreme case where \textit{all} objects are self-pasted, there is only a very minor drop in accuracy.

\begin{table} [t]
\centering
\small
\begin{tabular}{l|c|cccccc}
& \multicolumn{7}{c}{Inpainting bbox expansion} \\
\midrule
Model & 5\% & 0\% & 10\% & 25\% & 50\% & 75\% & 100\% \\
\midrule
\textsc{Qwen3-VL 4B Instruct} & 24.7 & 25.9 & 26.3 & 27.6 & 31.9 & 35.8 & 37.6 \\
\textsc{Qwen3-VL 8B Instruct} & 27.6 & 27.8 & 27.8 & 28.8 & 34.8 & 39.5 & 44.1 \\
\textsc{Gemini 2.5 Pro (MR)}  & 29.4 & 30.3 & 28.2 & 29.1 & 39.5 & 43.2 & 44.9 \\
\textsc{GPT-5 (LR)}  & 34.2 & 31.7 & 30.6 & 34.0 & 43.3 & 42.8 & 48.0 \\

\bottomrule

\end{tabular}

 \caption{\textbf{Inpainting artifacts are too small to confound evaluation}. We gradually increase the bounding box for inpainting until the artifacts are too large to notice (which occurs around 25-50\%). We use 5\% when generating pairs, which is far smaller than this threshold. }
 \vspace{-0.25cm}
\label{tab:eval-fraction}

\end{table}
Second, we ask at what threshold the artifacts actually start to make a difference on the model, and how close we are to that limit. Unlike self-pasting experiments where we hope for the accuracy to stay invariant to changes, this is a \textit{destructive study} where we actively increase the inpainting artifacts until we see substantial increases in accuracy because of confounding artifacts. In Table \ref{tab:eval-fraction}, we observe that our bounding box expansion of 5\% does not create artifacts large enough for the model to notice. If we do no expansion (0\%), this can sometimes make it easier to spot as the cut-and-paste will not appear natural. When increasing the inpaint area we only start to see minor increases at around 25\% expansion and it is not until 50\% expansion till the increase becomes prominent. These results (along with the previous self-pasting experiments) suggest that an increase in accuracy on our benchmark implies an improved 3D understanding and is not due to confounding artifacts. 

\section{Conclusion}

We introduced \textit{spatial inconsistencies}, a form of evaluating MLLMs' 3D understanding by asking them to spot what is incorrect in a geometrically impossible scene. We show how to synthesize them using a simple and scalable cut-and-paste method and use them to evaluate MLLMs. Our results demonstrate that while this task is simple for humans, it remains challenging for MLLMs. Furthermore, we conduct an in-depth investigation of our results and find that MLLMs exhibit large variance across scenes, objects, and their attributes, indicating their 3D understanding is very brittle and inconsistent.  We hope our work inspires others to look beyond training multimodal models to simply describe the visual world, and instead to teach them its fundamental physical and spatial constraints.

\bibliography{colm2026_conference}
\bibliographystyle{colm2026_conference}

\appendix
\clearpage
\section{Extended Results}
We had several figures evaluating our best models on various splits of the dataset. Here, we report accuracy on these same splits for all models. 
\begin{table}[h]
\centering
\resizebox{\textwidth}{!}{%
\begin{tabular}{l | ccc | ccc | cc}
\toprule
\textbf{Model} &
\multicolumn{3}{c}{\textbf{Object Depth}} &
\multicolumn{3}{c}{\textbf{Scene Lighting}} &
\multicolumn{2}{c}{\textbf{Plausibility}} \\
& 
\textbf{Close} & \textbf{Medium} & \textbf{Far} &
\textbf{Dark} & \textbf{Medium} & \textbf{Bright} &
\textbf{IP} & \textbf{P} \\
\midrule
Random Chance                          & 7.2  & 8.0  & 8.4  & 7.7  & 8.1  & 7.8  & 7.9  & 7.7  \\
Human                                  & 83.3 & 87.3 & 83.8 & 82.3 & 85.4 & 86.8 & 86.3 & 80.1 \\
\midrule
Gemma 3 12B                            & 13.2 & 7.3  & 4.9  & 7.8  & 11.2 & 6.3  & 9.4  & 5.4  \\
Idefics3 8B                            & 8.8  & 10.7 & 7.3  & 8.8  & 9.8  & 8.3  & 7.7  & 12.8 \\
Idefics2 8B                            & 9.8  & 12.2 & 13.7 & 12.2 & 13.2 & 10.2 & 12.4 & 10.1 \\
GPT-5 Nano                             & 14.6 & 17.6 & 13.7 & 16.1 & 15.1 & 14.6 & 16.1 & 12.8 \\
Qwen2.5 VL 7B                          & 11.7 & 12.7 & 23.4 & 19.0 & 19.0 & 9.8  & 17.1 & 12.2 \\
InternVL 3.5 8B                        & 14.2 & 13.7 & 20.5 & 16.6 & 16.1 & 15.6 & 16.5 & 14.9 \\
LLaVA OneVision 1.5 8B                 & 11.2 & 14.6 & 23.9 & 18.1 & 19.0 & 12.7 & 17.3 & 14.2 \\
Gemini 2.5 Flash                       & 18.1 & 18.1 & 16.6 & 18.1 & 21.0 & 13.7 & 18.4 & 14.9 \\
SpaceQwen2.5 VL 3B                     & 23.4 & 16.6 & 14.6 & 20.5 & 19.5 & 14.6 & 19.3 & 14.9 \\
GPT-4o                                 & 16.1 & 17.1 & 23.9 & 22.9 & 14.2 & 20.0 & 19.9 & 16.2 \\
Llama 3.2 Multimodal 11B               & 30.7 & 23.4 & 16.1 & 21.0 & 25.4 & 23.9 & 24.2 & 21.0 \\
Qwen3 VL 4B                            & 31.2 & 24.9 & 18.1 & 23.9 & 25.4 & 24.9 & 25.9 & 21.0 \\
Qwen3 VL 8B Thinking                   & 18.1 & 24.4 & 33.2 & 31.2 & 27.8 & 16.6 & 26.8 & 20.3 \\
Qwen3 VL 8B Instruct                   & 26.8 & 25.4 & 30.7 & 27.8 & 32.2 & 22.9 & 29.8 & 21.0 \\
Gemini 2.5 Pro (HR)                    & 28.3 & 30.2 & 28.3 & 29.8 & 33.7 & 23.4 & 29.1 & 28.4 \\
Gemini 2.5 Pro (LR)                    & \textbf{31.7} & 29.3 & 27.3 & 30.2 & 31.7 & 26.3 & 30.6 & 25.7 \\
Gemini 2.5 Pro (MR)                    & 30.7 & \textbf{32.2} & 25.4 & 30.7 & 32.7 & 24.9 & 29.8 & 28.4 \\
GPT-5 (HR)                             & 28.3 & 25.9 & 36.6 & 28.8 & 32.2 & 29.8 & 30.6 & 29.1 \\
GPT-5 (MR)                             & 23.4 & 30.2 & 40.5 & \textbf{32.2} & 33.2 & 28.8 & 33.2 & 25.7 \\
GPT-5 (LR)                             & 27.3 & 29.8 & \textbf{45.4} & 30.2 & \textbf{36.6} & \textbf{35.6} & \textbf{35.3} & \textbf{30.4} \\
\bottomrule
\end{tabular}}
\vspace{.1cm}
\caption{
\textbf{Performance breakdown (\% accuracy) by depth, lighting, and physical plausibility.} We report the full results for each model, of which a subset was shown in Figure \ref{fig:depth_light_plaus_num_label_acc}.
}
\label{tab:breakdown}
\end{table}


\clearpage
\begin{table*}[htbp]
\centering
\scriptsize

\resizebox{\textwidth}{!}{%
\begin{tabular}{l | cccccc}
\toprule
\textbf{Model} & \textbf{Chair} & \textbf{Window} & \textbf{Picture} & \textbf{Sofa} & \textbf{Cabinet} & \textbf{Door} \\
\midrule
Random Chance & 8.4 & 7.0 & 7.4 & 8.3 & 7.0 & 7.6 \\
Human & 86.8 & 82.5 & 86.4 & 84.8 & 88.1 & 85.7 \\
\midrule
Gemma 3 12B & 15.3 & 4.1 & 8.1 & 7.3 & 8.3 & 11.4 \\
Idefics3 8B & 8.5 & 1.4 & 4.8 & 17.1 & 11.1 & 11.4 \\
Idefics2 8B & 4.2 & 19.2 & 19.4 & 7.3 & 13.9 & 5.7 \\
GPT-5 Nano & 13.6 & 4.1 & 19.4 & 12.2 & 11.1 & 22.9 \\
Qwen2.5 VL 7B & 7.6 & 20.5 & 24.2 & 4.9 & 16.7 & 31.4 \\
InternVL 3.5 8B & 2.5 & 20.5 & 30.6 & 0.0 & 19.4 & 37.1 \\
LLaVA OneVision 1.5 8B & 15.3 & 11.0 & 19.4 & 12.2 & 19.4 & 31.4 \\
Gemini 2.5 Flash & 23.7 & 2.7 & 17.7 & 9.8 & 36.1 & 14.3 \\
SpaceQwen2.5 VL 3B & 22.0 & 1.4 & 12.9 & \textbf{39.0} & 2.8 & 14.3 \\
GPT-4o & 22.0 & 13.7 & 22.6 & 12.2 & 16.7 & 31.4 \\
Llama 3.2 Multimodal 11B & 28.8 & 13.7 & 27.4 & 19.5 & 16.7 & 17.1 \\
Qwen3 VL 4B & 32.2 & 5.5 & 29.0 & 26.8 & 11.1 & 31.4 \\
Qwen3 VL 8B Thinking & 18.6 & 17.8 & 37.1 & 17.1 & 33.3 & 54.3 \\
Qwen3 VL 8B Instruct & 16.1 & 17.8 & 56.5 & 12.2 & 25.0 & 37.1 \\
Gemini 2.5 Pro (HR) & 35.6 & 11.0 & 40.3 & 34.1 & 33.3 & 20.0 \\
Gemini 2.5 Pro (LR) & \textbf{39.8} & 9.6 & 25.8 & 31.7 & 33.3 & 20.0 \\
Gemini 2.5 Pro (MR) & \textbf{39.8} & 9.6 & 24.2 & 29.3 & \textbf{38.9} & 20.0 \\
GPT-5 (HR) & 22.0 & 24.7 & 41.9 & 9.8 & 36.1 & 42.9 \\
GPT-5 (MR) & 19.5 & \textbf{28.8} & 51.6 & 17.1 & \textbf{38.9} & \textbf{57.1} \\
GPT-5 (LR) & 23.7 & 26.0 & \textbf{61.3} & 14.6 & \textbf{38.9} & 51.4 \\
\bottomrule
\end{tabular}%
}

\vspace{0.75em}

\resizebox{\textwidth}{!}{%
\begin{tabular}{l | ccccccc}
\toprule
\textbf{Model} & \textbf{Lamp} & \textbf{Mirror} & \textbf{Television} & \textbf{Floormat} & \textbf{Blinds} & \textbf{Table} & \textbf{Misc.} \\
\midrule
Random Chance & 5.4 & 11.7 & 9.1 & 6.7 & 9.2 & 6.6 & 7.9 \\
Human & 85.7 & 78.1 & 84.6 & 86.4 & 94.4 & 66.7 & 83.6 \\
\midrule
Gemma 3 12B & 3.7 & 4.0 & 0.0 & 10.5 & 0.0 & 0.0 & 9.3 \\
Idefics3 8B & 3.7 & 8.0 & 4.5 & 5.3 & 6.7 & 15.4 & 14.0 \\
Idefics2 8B & 25.9 & 28.0 & 18.2 & 5.3 & 20.0 & 0.0 & 7.8 \\
GPT-5 Nano & 11.1 & 20.0 & 22.7 & 15.8 & 0.0 & 0.0 & 23.3 \\
Qwen2.5 VL 7B & 18.5 & 24.0 & 22.7 & 0.0 & 6.7 & 0.0 & 17.8 \\
InternVL 3.5 8B & 51.9 & 20.0 & 22.7 & 0.0 & 6.7 & 0.0 & 13.2 \\
LLaVA OneVision 1.5 8B & 14.8 & 16.0 & 27.3 & 0.0 & 6.7 & 7.7 & 19.4 \\
Gemini 2.5 Flash & 7.4 & 24.0 & 13.6 & 10.5 & 13.3 & 15.4 & 21.7 \\
SpaceQwen2.5 VL 3B & 44.4 & 8.0 & 22.7 & 5.3 & 13.3 & \textbf{23.1} & 23.3 \\
GPT-4o & 3.7 & 24.0 & 27.3 & 15.8 & 6.7 & 7.7 & 20.9 \\
Llama 3.2 Multimodal 11B & 59.3 & 20.0 & 18.2 & 10.5 & 13.3 & 7.7 & 25.6 \\
Qwen3 VL 4B & 40.7 & 28.0 & 40.9 & 10.5 & 6.7 & 7.7 & 27.1 \\
Qwen3 VL 8B Thinking & 18.5 & \textbf{56.0} & 31.8 & 0.0 & 6.7 & 0.0 & 24.8 \\
Qwen3 VL 8B Instruct & \textbf{66.7} & 52.0 & 36.4 & 0.0 & 13.3 & 0.0 & 27.1 \\
Gemini 2.5 Pro (HR) & 18.5 & 16.0 & 13.6 & \textbf{26.3} & 6.7 & 7.7 & 39.5 \\
Gemini 2.5 Pro (LR) & 22.2 & 24.0 & 22.7 & \textbf{26.3} & 20.0 & 7.7 & \textbf{41.1} \\
Gemini 2.5 Pro (MR) & 25.9 & 28.0 & 18.2 & 21.1 & 20.0 & 15.4 & 40.3 \\
GPT-5 (HR) & 55.6 & 28.0 & 40.9 & 5.3 & \textbf{40.0} & 15.4 & 34.1 \\
GPT-5 (MR) & 33.3 & 40.0 & \textbf{54.5} & 0.0 & 6.7 & 0.0 & 34.1 \\
GPT-5 (LR) & 33.3 & 48.0 & 40.9 & \textbf{26.3} & \textbf{40.0} & 7.7 & 34.9 \\
\bottomrule
\end{tabular}%
}

\caption{\textbf{Accuracy by inconsistent object class.} We report the full results for each model, of which a subset was shown in Figure \ref{fig:object_scene_class}.}
\label{tab:object-class}
\end{table*}
\clearpage

\begin{table*}[htbp]
\centering
\scriptsize

\resizebox{\textwidth}{!}{%

\begin{tabular}{l | ccccc}
\toprule
\textbf{Model} & \textbf{Lounge} & \textbf{Kitchen} & \textbf{Study} & \textbf{Hallway} & \textbf{Bedroom} \\
\midrule
Random Chance & 8.4 & 6.3 & 7.1 & 8.7 & 8.1 \\
Human & 88.9 & 83.3 & 81.0 & 85.9 & 85.7 \\
\midrule
Gemma 3 12B & 5.6 & 8.5 & 17.4 & 9.0 & 4.8 \\
Idefics3 8B & 8.8 & 6.8 & 9.2 & 9.0 & 7.9 \\
Idefics2 8B & 12.0 & 9.4 & 16.5 & 7.5 & 11.1 \\
GPT-5 Nano & 9.6 & 17.1 & 21.1 & 11.9 & 12.7 \\
Qwen2.5 VL 7B & 12.8 & 18.8 & 20.2 & 17.9 & 15.9 \\
InternVL 3.5 8B & 13.6 & 13.7 & 27.5 & 14.9 & 12.7 \\
LLaVA OneVision 1.5 8B & 19.2 & 14.5 & 14.7 & 23.9 & 15.9 \\
Gemini 2.5 Flash & 18.4 & 12.8 & 15.6 & 19.4 & 17.5 \\
SpaceQwen2.5 VL 3B & 19.2 & 12.8 & 16.5 & 23.9 & 12.7 \\
GPT-4o & 22.4 & 12.8 & 19.3 & 19.4 & 28.6 \\
Llama 3.2 Multimodal 11B & 25.6 & 19.7 & 24.8 & 17.9 & 31.7 \\
Qwen3 VL 4B & 20.8 & 18.8 & 26.6 & 19.4 & 28.6 \\
Qwen3 VL 8B Thinking & 19.2 & 22.2 & 33.0 & \textbf{31.3} & 20.6 \\
Qwen3 VL 8B Instruct & 18.4 & 30.8 & 30.3 & 25.4 & 28.6 \\
Gemini 2.5 Pro (HR) & 33.6 & 23.9 & 33.9 & \textbf{31.3} & 23.8 \\
Gemini 2.5 Pro (LR) & 32.8 & 21.4 & \textbf{36.7} & \textbf{31.3} & 22.2 \\
Gemini 2.5 Pro (MR) & \textbf{36.0} & 23.9 & 28.4 & 29.9 & 27.0 \\
GPT-5 (HR) & 31.2 & 25.6 & 33.9 & 25.4 & 31.7 \\
GPT-5 (MR) & 28.8 & \textbf{32.5} & 31.2 & 29.9 & 31.7 \\
GPT-5 (LR) & 32.0 & \textbf{32.5} & 33.0 & 29.9 & \textbf{47.6} \\
\bottomrule
\toprule
\textbf{Model} & \textbf{Bathroom} & \textbf{Patio} & \textbf{Restaurant} & \textbf{Entertainment} & \textbf{Misc} \\
\midrule
Random Chance & 10.3 & 7.5 & 4.8 & 11.2 & 10.6 \\
Human & 83.9 & 87.5 & 80.0 & 82.4 & 80.0 \\
\midrule
Gemma 3 12B & 6.7 & 2.9 & 4.0 & 6.7 & 7.1 \\
Idefics3 8B & 13.3 & 8.6 & 8.0 & 13.3 & 14.3 \\
Idefics2 8B & 17.8 & 8.6 & 4.0 & 26.7 & 7.1 \\
GPT-5 Nano & 22.2 & 17.1 & 12.0 & 6.7 & 21.4 \\
Qwen2.5 VL 7B & 13.3 & 5.7 & 20.0 & 20.0 & 0.0 \\
InternVL 3.5 8B & 11.1 & 8.6 & 24.0 & 20.0 & 7.1 \\
LLaVA OneVision 1.5 8B & 13.3 & 8.6 & 12.0 & 20.0 & 28.6 \\
Gemini 2.5 Flash & 20.0 & 22.9 & 20.0 & 26.7 & 21.4 \\
SpaceQwen2.5 VL 3B & 17.8 & 28.6 & 28.0 & 20.0 & 21.4 \\
GPT-4o & 17.8 & 17.1 & 4.0 & 20.0 & 28.6 \\
Llama 3.2 Multimodal 11B & 24.4 & 14.3 & 32.0 & 20.0 & 21.4 \\
Qwen3 VL 4B & 26.7 & 22.9 & \textbf{48.0} & 40.0 & 42.9 \\
Qwen3 VL 8B Thinking & 28.9 & 14.3 & 28.0 & 33.3 & 35.7 \\
Qwen3 VL 8B Instruct & 28.9 & 25.7 & 36.0 & 33.3 & \textbf{50.0} \\
Gemini 2.5 Pro (HR) & 33.3 & 20.0 & 20.0 & 20.0 & 35.7 \\
Gemini 2.5 Pro (LR) & \textbf{35.6} & 22.9 & 20.0 & 26.7 & \textbf{50.0} \\
Gemini 2.5 Pro (MR) & \textbf{35.6} & 20.0 & 28.0 & 20.0 & \textbf{50.0} \\
GPT-5 (HR) & 26.7 & 37.1 & 28.0 & 33.3 & 42.9 \\
GPT-5 (MR) & \textbf{35.6} & 34.3 & 20.0 & \textbf{46.7} & 35.7 \\
GPT-5 (LR) & 31.1 & \textbf{40.0} & 32.0 & 26.7 & 42.9 \\
\bottomrule

\end{tabular}%
}
\caption{\textbf{Accuracy by scene class.} We report the full results for each model, of which a subset was shown in Figure \ref{fig:object_scene_class}.}

\label{tab:scene-class}
\end{table*}
\clearpage

\begin{table*}[htbp]
\centering

\begin{tabular}{l | cccc}
\toprule
\textbf{Model} & \textbf{5--10} & \textbf{11--15} & \textbf{16--20} & \textbf{21--26} \\
\midrule
Random Chance & 12.3 & 7.9 & 5.7 & 4.1 \\
Human & 86.5 & 86.8 & 85.6 & 78.5 \\
\midrule
Gemma 3 12B & 12.1 & 6.5 & 7.3 & 7.5 \\
Idefics3 8B & 13.3 & 9.7 & 6.5 & 4.5 \\
Idefics2 8B & 16.2 & 9.2 & 11.4 & 10.4 \\
GPT-5 Nano & 19.1 & 13.0 & 18.7 & 10.4 \\
Qwen2.5 VL 7B & 21.4 & 15.7 & 13.0 & 11.9 \\
InternVL 3.5 8B & 16.8 & 12.4 & 13.0 & 23.1 \\
LLaVA OneVision 1.5 8B & 22.0 & 14.1 & 13.8 & 15.7 \\
Gemini 2.5 Flash & 22.5 & 17.8 & 14.6 & 13.4 \\
SpaceQwen2.5 VL 3B & 24.9 & 16.2 & 10.6 & 19.4 \\
GPT-4o & 27.2 & 20.5 & 17.9 & 7.5 \\
Llama 3.2 Multimodal 11B & 28.3 & 17.8 & 20.3 & 27.6 \\
Qwen3 VL 4B & 30.6 & 18.4 & 24.4 & 26.1 \\
Qwen3 VL 8B Thinking & 31.8 & 25.9 & 20.3 & 20.1 \\
Qwen3 VL 8B Instruct & 31.8 & 25.4 & 25.2 & 27.6 \\
Gemini 2.5 Pro (HR) & 28.9 & 31.9 & 30.1 & 23.9 \\
Gemini 2.5 Pro (LR) & 32.9 & 34.1 & 26.0 & 21.6 \\
Gemini 2.5 Pro (MR) & 34.1 & 33.5 & 23.6 & 23.1 \\
GPT-5 (HR) & 34.1 & 30.8 & 26.8 & 27.6 \\
GPT-5 (MR) & \textbf{38.2} & 32.4 & 28.5 & 23.9 \\
GPT-5 (LR) & 37.0 & \textbf{35.1} & \textbf{31.7} & \textbf{31.3} \\
\bottomrule
\end{tabular}%
\caption{\textbf{Accuracy by number of labels.} We report the full results for each model, of which a subset was shown in Figure \ref{fig:depth_light_plaus_num_label_acc}.}
\label{tab:num-labels}
\end{table*}
\section{Applications of Spatial Inconsistencies}

\begin{table}
  \centering
  \begin{tabular}{l|c}
    \toprule
    Model & Accuracy on DL3DV inconsistencies\\
    \midrule
    Random Chance & 18.4 \\
    \midrule
    Qwen3-VL 4B & 47.7 \\ 
    Qwen3-VL 8B & 57.7 \\
    Gemini 2.5 Pro (MR) & 52.9 \\
    GPT-5 (LR) & 52.4 \\
    \bottomrule
  \end{tabular} \vspace{0.1cm}
  \caption{When we scale up to real-world data, our task is still challenging for models. }
  \label{dl3dv_acc}
\end{table}

\subsection{Scaling Up to Real-World Data}
We scale up the dataset by generating pairs using DL3DV dataset, which contains over 10K real-world multiview scenes. We obtain object tracks using SAM 3 (captioned by Qwen3-VL 8B) and metric depths with Depth Pro. We generated {\bf 39K pairs} from over {\bf 6,900 scenes} with diverse settings such as parks, malls, gyms and more. We show the results of the scaled up dataset (DL3DV) in Table \ref{dl3dv_acc}. We observe that this task is still challenging for MLLMs. We see that MLLMs score higher than the Hypersim set, which we assume is due to the substantially lower number of objects per scene. This is both due to the expensive cost of SAM3 for tracking and DL3DV's selection of outdoor scenes, which have less objects. We evaluated Qwen models with the full 39K set, but had to use a fixed random sample of 6K for Gemini/GPT-5 due to API limits and cost. 
\subsection{Does Learning to Spot Inconsistencies Improve Broader 3D Understanding?}
Given the poor zero-shot accuracy of state-of-the-art multimodal models, a natural question arises: is learning to spot spatial inconsistencies a "useful" task? We draw on ideas from visual representation learning and use our generated inconsistencies as a pretext task. If our task is sufficiently related to downstream 3D understanding tasks, we expect to see a small but present improvement on zero-shot accuracy over the base model. To answer this question, we fine-tune Qwen3-4B using our autogenerated set of 10K pairs. We apply LoRA~\citep{hu2021loralowrankadaptationlarge} with rank $16$. For Table~\ref{tab:blink}, we train for one epoch using our scaled-up inconsistency sets.  

\begin{table}
\centering
\small
\scalebox{0.96}{
\begin{tabular}{lccc}
\toprule
\textbf{Model} & \textbf{Base Model} & \textbf{Hypersim}  & \textbf{DL3DV}\\
\midrule
\textcolor{gray}{Hypersim Inconsistencies}  & \textcolor{gray}{24.7} & \textcolor{gray}{66.8$\pm$1.5} & \textcolor{gray}{55.2} \\
Multiview Reasoning                 & 37.6 & 39.9$\pm$1.4 & 43.6 \\
Visual Correspondence                 & 80.2 & 81.7$\pm$0.7 & 82.2 \\
Relative Depth                 & 84.7 & 86.9$\pm$1.1 & 87.7\\

\bottomrule
  \end{tabular} }
\vspace{0.1cm}
\caption{We show that Qwen3-VL 4B finetuned with additional scene diversity perform on par or better than our initial models at 3D perception and perform well on the Hypersim evaluation set. }
\vspace{-.1in}
\label{tab:blink}
\end{table}

\paragraph{Transfer to broader 3D benchmarks.} 
The more important question is whether this training transfers beyond our task. We evaluate the fine-tuned model on the BLINK benchmark~\citep{fu2024blink}, focusing on categories most closely tied to spatial geometry: ``Multiview Reasoning'', ``Visual Correspondence'', and ``Relative Depth''. As summarized in Table~\ref{tab:blink}, training on our inconsistency-spotting task yields consistent improvements across all three metrics. We observe gains of approximately $2.3$ percentage points in both ``Multiview Reasoning'' and ``Relative Depth'', and $1.3$ percentage points in ``Visual Correspondence''. We also finetune our model on our scaled-up DL3DV dataset and see similar gains. These improvements are notable because the model receives no direct supervision on depth estimation, correspondence matching, or camera motion. Instead, identifying \emph{which} object breaks cross-view consistency implicitly encourages the model to learn depth (for scale compatibility), correspondence (for appearance matching), and also camera motion understanding. These findings suggest our simple cut-and-paste approach does provide signal to learn 3D understanding from. 

\paragraph{In-domain generalization.} 
We also retrain the model with 20\% of scenes held-out to ensure that training on our task can generalize to novel scenes. As shown in Table~\ref{tab:disjoint}, fine-tuning yields a substantial improvement in performance: Qwen3 VL-4B jumps from $24.7\%$ to $69.2\%$ accuracy on entirely unseen scenes. This signals that the task is learnable and that our synthetic pairs provide a consistent and meaningful signal for the model to acquire a generalizable notion of 3D violations, rather than memorizing scene layouts or object textures.  

\begin{table}[!t]
\centering
\small
\begin{tabular}{lcc}
\toprule
\textbf{Model} & \textbf{Base Model} & \textbf{Finetuned}\\
\midrule
Our Benchmark, disjoint set                & 24.7 & 69.2 \\

\bottomrule
  \end{tabular} \vspace{0.1cm}

\caption{
\textbf{Fine-tuning on our spatial inconsistency task yields large gains on a disjoint test split.}
Accuracy (\%) of the base vision–language model versus the same model fine-tuned on our benchmark, evaluated on a held-out set of image pairs with no scene overlap. Performance jumps from 24.7\% to 69.2\%, showing that the task is learnable and that targeted training substantially improves spatial reasoning on new scenes. 
}
\label{tab:disjoint}
\end{table}

\subsection{Do VQA Judges Notice Spatial Inconsistencies?}

Despite struggling at directly identifying inconsistent objects,
modern MLLMs are widely used as VQA-style judges for evaluating
image and video generation. We therefore ask: do these judges
notice our 3D inconsistencies when used in their standard Yes/No
evaluation protocol? Given a set of frames and a textual description, the model is asked a constrained question such as: 
\textit{``Does this video show \{\texttt{caption}\}? Please answer Yes or No.''} The alignment score is then estimated from the probability assigned to the \texttt{Yes} token. In practice, this setup is used not only to measure semantic alignment, but also as a proxy for visual plausibility: if two candidates match the prompt, the expectation is that the judge will assign a higher \texttt{Yes} score to the one that is more realistic.

Recent benchmarks (e.g., GenAI-Bench \citep{li2024genaibench}) support this intuition by showing that VQA-style judges can successfully rank images that all satisfy the prompt but differ in fidelity and artifacts, suggesting some sensitivity to realism beyond pure text alignment. We hypothesize that this sensitivity to realism does \textit{not} extend to 3D understanding, and thus may be unable to distinguish videos where object poses are modified such that the frames are impossible in a static scene.

\begin{table}[!t]
\centering
\small
\begin{tabular}{lccc}
\toprule
\textbf{Model} & \textbf{Pairwise} & \textbf{Pearson} & \textbf{Kendall} \\
\midrule
GPT\textendash4o                 & 50.1 & -0.14 & 0.32 \\
\bottomrule
  \end{tabular} \vspace{0.1cm}

\caption{
\textbf{VQAScore evaluation of GPT\textendash4o on our spatial inconsistency benchmark.}
We report the pairwise accuracy (\%) of VQAScore \citep{lin2024evaluating} when asked to score both the physically consistent image and the manipulated one, along with Pearson and Kendall correlations between VQAScore and human accuracy across image pairs. The near-chance pairwise score and weak correlations indicate that the SOTA VQA-based judge poorly aligns with human assessments of spatial consistency.
}
\label{tab:vqascore}
\end{table} 
We use our constructed pairs to test this assumption directly. For each multiview set $(V_1, V_2, V_2')$:
\begin{enumerate}
\item We first generate a caption $C_1$ from frame $V_1$ alone (without labels) using GPT-5.
\item We treat $(V_1, V_2)$ and $(V_1, V_2')$ as the frames sampled from two candidate videos depicting the same static scene. 
\item We compute the accuracy as the fraction of items where the unmodified pair $(V_1,V_2)$ receives a higher \texttt{Yes} score than the manipulated pair $(V_1,V_2')$.
\end{enumerate} 
Assuming GenAI-Bench's finding holds, a judge that truly understands scene geometry should consistently prefer $(V_1, V_2)$, the unmodified pair over $(V_1, V_2')$ which is spatially inconsistent. However, as shown in Table~\ref{tab:vqascore}, GPT-4o (the current SOTA for VQAScore eval) performs near random chance. This suggests that VQA judges does not spot spatial inconsistencies when ranking alignment of generated videos and provided captions. Additionally, when utilizing VQA judges to \textit{control} diffusion-based generators \citep{luo2025dualprocess, zheng2026controlling}, they may amplify these spatial inconsistencies due to their inability to identify them. 

\end{document}